\documentclass[10pt,letterpaper]{article}

\usepackage{cogsci}

\cogscifinalcopy % Uncomment this line for the final submission 

\usepackage{graphicx}
\usepackage{algpseudocode}
\usepackage{algorithm}
\usepackage{algorithmicx}

\usepackage{pslatex}
\usepackage{apacite}
\usepackage{float} % Roger Levy added this and changed figure/table
\usepackage{tabularx}
\usepackage{makecell}
\usepackage{multirow}
\usepackage{booktabs} 
\title{Learning to Play Video Games with Intuitive Physics Priors}
 
\author{{\large \bf Abhishek Jaiswal (abhijais@cse.iitk.ac.in)} \\
  Department of CSE, IIT Kanpur \\
  Kalyanpur, UP 208016 India 
  \AND {\large \bf Nisheeth Srivastava (nsrivast@cse.iitk.ac.in)} \\
Department of CSE, IIT Kanpur \\
  Kalyanpur, UP 208016 India }

\begin{document}

\maketitle

\begin{abstract}
Video game playing is an extremely structured domain where algorithmic decision-making can be tested without adverse real-world consequences. While prevailing methods rely on image inputs to avoid the problem of hand-crafting state space representations, this approach systematically diverges from the way humans actually learn to play games. In this paper, we design object-based input representations that generalize well across a number of video games. Using these representations, we evaluate an agent's ability to learn games similar to an infant - with limited world experience, employing simple inductive biases derived from intuitive representations of physics from the real world. Using such biases, we construct an object category representation to be used by a Q-learning algorithm and assess how well it learns to play multiple games based on observed object affordances. Our results suggest that a human-like object interaction setup capably learns to play several video games, and demonstrates superior generalizability, particularly for unfamiliar objects. Further exploring such methods will allow machines to learn in a human-centric way, thus incorporating more human-like learning benefits.

\end{abstract}

\section{Introduction}
% \textbf{Questions - Flowchart or algorithm or both,\\ state representatation image needed?\\ above limit by 1 page\\}

%Video games bring in a rich intermix of interacting physical systems, capturing real-world dynamics in a tractable fashion. 
%Consequently, game-playing algorithms have received substantial attention from the reinforcement learning community as they strive to decipher the intricacies of human-like learning."
%Nevertheless, amidst the rapid advancement of deep learning and Big Data capabilities, the emphasis has streamed towards feeding models with vast datasets, with multiple stability enforcing techniques, in a model-free fashion to beat the state-of-the-art results.

Deep reinforcement learning (DRL) algorithms have shown professional to superhuman competency in gaming environments such as MuJoCo, and Atari \cite{shakya2023reinforcement, goodfellow2014explaining}. But, at the same time, like other black box deep learning models, they can break with even slight modifications of the environment~\cite {justesen2018illuminating,goodfellow2014explaining}.

\begin{figure}[tbhp]
    \centering
    \includegraphics[width=.95\linewidth]{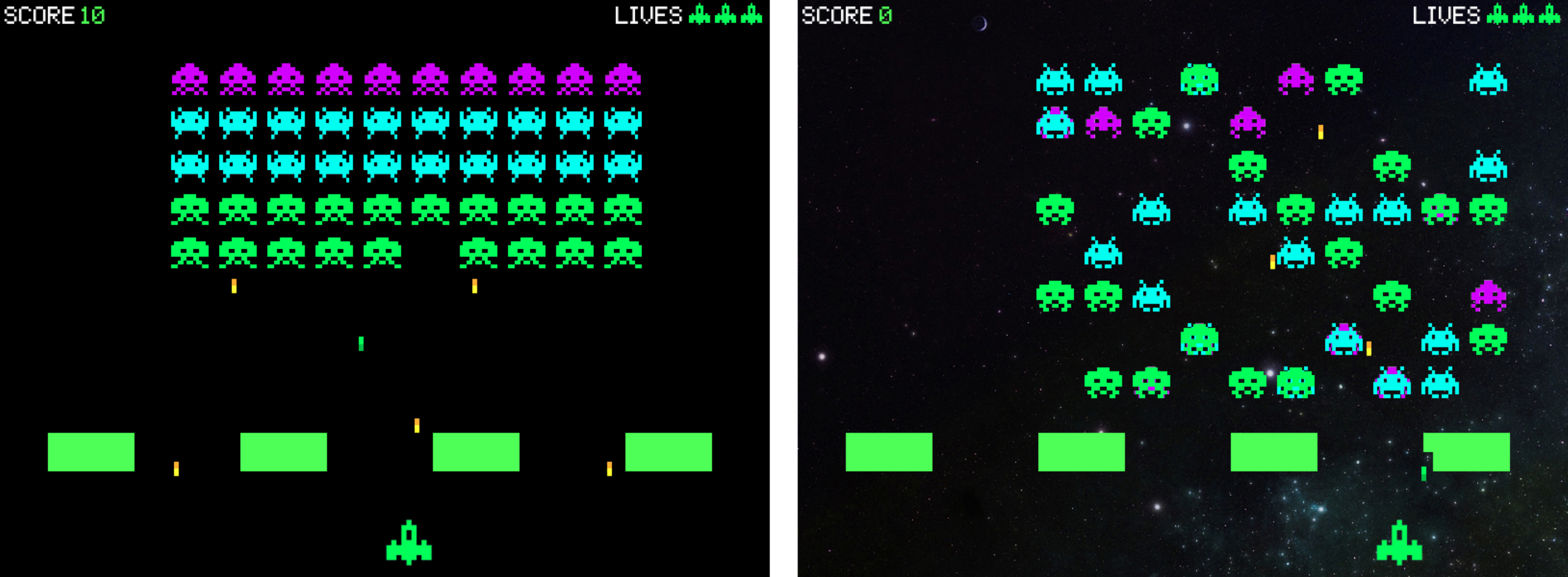}
    \vspace{-2.5mm}
    \caption{Simple Variations, Crippling Results - Deep Learning Models break even with a slight variation of the environment (Right image - partially randomized enemy positions).}
    \label{fig:drl_adversary}
\end{figure}

For example, to contrast human and machine-level learning, Figure \ref{fig:drl_adversary} shows two variants of the space invaders game we tested. DQN was trained on the basic version on the left for one million iterations and tested on the variant with partially randomized enemy positions on the right. The base variant's average score was 510, whereas the right variant could score only 280; right around random performance. On the contrary, humans play through such variants with ease. Also, DRL-based approaches still fail at generalizing and transferring learned knowledge to novel domains \cite{kansky2017schema}. Humans demonstrate superior learning trajectories, learning games quickly and also performing well on modifications \cite{tsividis2017human}.

%By conducting such simple experiments, we can sense the level of disparity between the learning approaches taken by humans and DRL models.

%Thus, this paper focuses on techniques to make machines learn in a more human-like fashion. The first distinction between the recent ML direction and the human way of learning is its input. DRL takes images as input which essentially means that the whole world for them is a collection of pixels. In contrast, when humans look at the world, they do not see pixels; instead, they see objects, which is the interest of the works on object-based reinforcement learning.

%Here also, we take a slightly different approach which we find more interesting. Instead of working directly with the objects and assuming their properties as given, we try to look at the game world from a fresh perspective, somewhat similar to the view of a child who does not come with an oversized baggage of existing knowledge. 

Attempting to bring the study of video game-playing closer to human cognitive behavior, in this paper, we learn game-playing using common human inductive biases. With this line of work, we aim to leverage the same advantages humans show in generalization and zero-shot transfer on related tasks. To this end, we design generalizable affordance-based representations of a reinforcement learning agent's state space using two primary assumptions. First, we try to incorporate the thinking of a first-time player in game playing using inductive biases drawn from humans' common core knowledge of intuitive physics~\cite{spelke1990principles,spelke2007core}. Second, we design a state space representation using object categories instead of classically used object-based input representations. We test the value of these representations by training a simple Q-learning agent~\cite{watkins1992q}, and comparing its performance against DQN~\cite{mnih2015human}. Finally, we show that rather than using standalone objects,  describing the game world in terms of object categories offers
% by using our approach of working with object categories rather than standalone objects, we see 
learning and generalization trends practically unattainable by resource-hungry pixel-based DRL agents.

\section{Using affordances to infer states}

% Not using this reference
% \cite{griffiths2010probabilistic}  \textbf{Since humans are capable of learning in many complex environments, one of the central questions of cognitive science is what kind of inductive biases empower us to do this}

%Deep Reinforcement Learning has been prolifically used in games, notably dominating the Atari Benchmarks and defeating professional players in games such as Go, Dota 2, and Chess, among others \cite{silver2018general,berner2019dota}.

%Like many deep learning methods, these algorithms work as a black box \cite{kumar2023disentangling}, often struggling with inexplicability and poor sample efficiency\cite{mohan2023structure}. More so, akin to their counterparts in deep learning \cite{goodfellow2014explaining}, they are susceptible to errors with even slight modifications of features~\cite{lu2020deep}. On the contrary, humans, against AI, despite being defeated on many of the gaming benchmarks, do much better at learning task abstractions to reuse the acquired knowledge \cite{kansky2017schema}.  

%Model-based Reinforcement Learning is better at sample efficiency. Model-based planning has also shown generalization behavior in complex tasks such as video games~\cite{pouncy2021model}. Out Of the many works on imparting human-like learning using models of the environment, impressive results have been achieved by Theory-based Reinforcement Learning methods.

Theory-based RL is a form of model-based reinforcement learning where the model is defined in terms of rich ontological symbolic representations pertaining to physical objects, their relations, and interactions. Using various intuitive theories, theory-based RL explicitly tries to incorporate human ways of learning~\cite{tsividis2021human}. Such intuitive theories stem from a core knowledge representation of the world visible even in infants who can segregate the visual input into ontological structures such as objects, goals, and physics~\cite{baillargeon2004infants,spelke1990principles,spelke2007core,csibra2008goal}. Humans also have been shown to make internal models using theory representation~\cite{tomov2023neural}. Similarly, semantic and syntactic biases, such as those used in theory-based RL, show a strong resemblance to human-like learning~\cite{pouncy2022inductive}. Humans show a wide range of flexibility in adapting to variations within the same task domain. As such,~ 
\citeA{pouncy2021model} have shown evidence that such flexibility, a hallmark of human intelligence,  can arise by representations composed of objects and interactions within a model-based framework. Thus, theory-based RL has shown a promising resemblance to human-like learning. However, being dependent on already possessing a detailed model of the environment, it has significant practical limitations.

%One of the footholds of theory-based Reinforcement Learning, which we also utilize, is the emphasis on structural representation in human cognition. Though contrary to previous works, instead of considering them given --- as a world model --- we learn the intricacies of the structural form.

Structure representation is considered a key ingredient to human-like learning. For example, \citeA{lake2017building} contrasted a set of key ingredients for more human-like learning against recent developments in deep reinforcement learning. \citeA{bapst2019structured} stressed structured inputs as a key to better generalization and solving situations beyond the training space. 
In a series of related works,~\shortciteA{doumas2022theory} show cross-game generalization by using relational analogies over symbolic representations, whereas we use categorial object affordance for solving game variants and hope to extend the concepts across games.

 \shortciteA{gershman2010learning} proposed using observation to infer states and actions through intuitive physics and intuitive psychology. 
To make sense of these observations, humans utilize various priors that help them explore efficiently. \citeA{dubey2018investigating} explore and quantify such priors for video gaming tasks. Our work builds upon such principles to learn a working structure of the world. 

\citeA{tsividis2021human} worked on the idea of making machines learn more like humans starting from early childhood state using strong theories about the working of the world. We take a slightly different approach and learn the affordances from object specifications rather than using pre-defined rules of interactions. Much like them, we also levy inductive biases for this task, which we understand to be a product of evolution, such as agent identification, threat perception, and goal attribution.

%In a related setting, using program induction,  \citeA{ellis2023dreamcoder} developed DreamCoder -- growing learning capabilities from a child-like state. Similarly, \citeA{ding2023embodied} used language instructions and human demonstrations to learn concepts, acting like a baby, learning from environmental interactions
%, as babies are known to learn by demonstrations \cite{meltzoff1999born}. 
 
%We explore this learning task by leveraging inductive biases and agent-object interactions with a focus on categorization, complementing, and yet differentiating with previous works.
 
The inductive biases we adopt are drawn primarily from the work of Elizabeth Spelke~\cite{spelke2007core}. \citeA{spelke1990principles} reason that infants perceive objects based on perceptual units moving together, moving separately, interacting on contact, and maintaining their shapes and sizes while in motion. We leverage these visual signals to learn fundamental affordances such as avoid, touch, and block through our Reinforcement Learning (RL) agent's actions trained exclusively on object-specific properties that are interpretable and in alignment with concepts of infant learning.

% In this line,\shortciteA{tsividis2017human} propose that humans form most of winning objective after observing game entity interaction which we also build upon by training a RL agent on object specific properties, letting the agent learn winning strategies from interactions in an interpretable way.

Thus, in this domain, akin to the work of \citeA{ding2023embodied} in the space of natural languages, we try to answer a simple question - `can we enable an agent to learn like a small child?' and test the hypothesis in game settings. 

Humans look at the world in terms of objects and their interactions; This is one of their core knowledge~\cite{spelke2007core}. Drawing on this insight, we shape the task of object reasoning around basic principles of core knowledge and show that we can achieve game-playing in a more cognitive and less mechanistic manner. Specifically, instead of just looking at objects in isolation, our agents infer meaningful object categories as part of their interactions with the game world. 

%This notion is inspired from compelling evidence on humans learning object categories in specific brain regions \cite{kriegeskorte2008matching, dicarlo2012does}. Then, we build upon this representation to learn category-level affordances and test our hypothesis on multiple tasks considered easy for humans but proven challenging for machines.

% \begin{figure*}[tbp]
% \centering
% \includegraphics[width=\linewidth]{Figs/CogSci-RoadRash.pdf}2
%     \caption{Space Invaders Original, Position Modification, Color and size Modification, Image Modification. For VGDL games all objects are of fixed shape and Size.\textbf{I will annotate categories in these images.} }
% \label{fig:gamesMod}
% \end{figure*}

% \begin{figure*}[tbp]
% \centering
%     \includegraphics[width=\linewidth]{Figs/CogSci-MyAliens.pdf}
%     \caption{Space Invaders Original, Position Modification, Color and size Modification, Image Modification. For VGDL games all objects are of fixed shape and Size.\textbf{I will annotate categories in these images.} }
% \label{fig:gamesMod}
% \end{figure*}
\section{Learning How to Play}

\begin{figure}[tbhp]
    \centering
    \includegraphics[width=.79\linewidth]{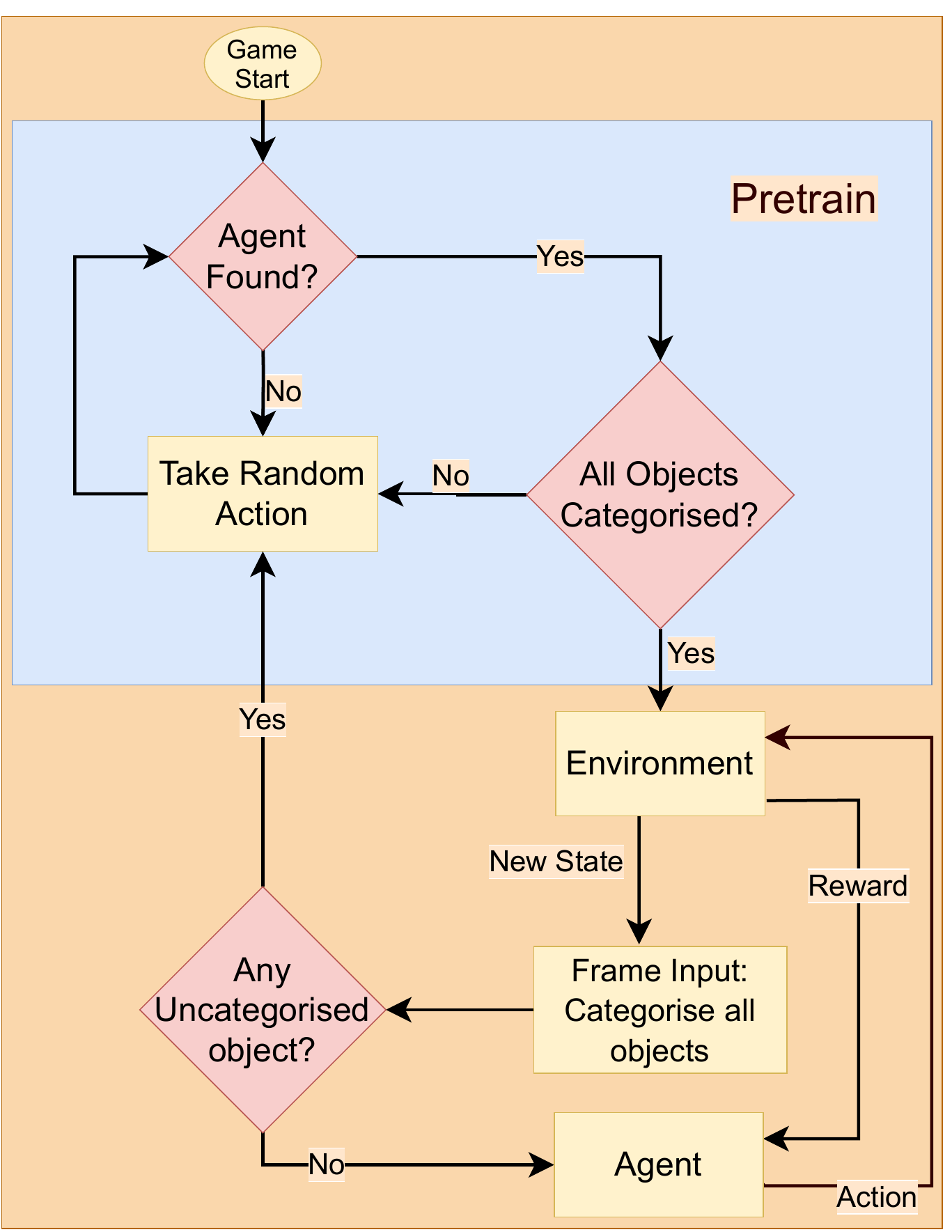}
        \vspace{-2.5mm}

    \caption{Schematic representation of Agent-Action pipeline based upon intuitive physics priors.}
    \label{fig:flowdiag}
\end{figure}

We look at the game screen from the view of a novice player holding a very minimal baggage of experience. Such players would see certain entities stand out on the screen by virtue of their specific forms, colors, or movements but would not know the associated affordances -- a task necessary to accomplish their desire to win \cite{csibra2008goal}. The first step in this learning process would be detecting what we control on the screen, i.e., the agent representing the player in the game. Agent detection is one of the key ideas in human-like learning and also a stark differentiator from large-scale machine-like pattern matching~\cite{de2023self}.
After knowing the where and how of the agent,
the next step would be to devise a locomotive strategy, necessitating knowledge of at least a minimal set of affordances associated with other game entities, for which we devise a set of representative object categories and learn category-level affordances . Refer to Figure \ref{fig:flowdiag} for an overview of the complete pipeline. %motivated from evoltion

\begin{figure*}[htbp]
\centering
\includegraphics[width=0.75\linewidth]{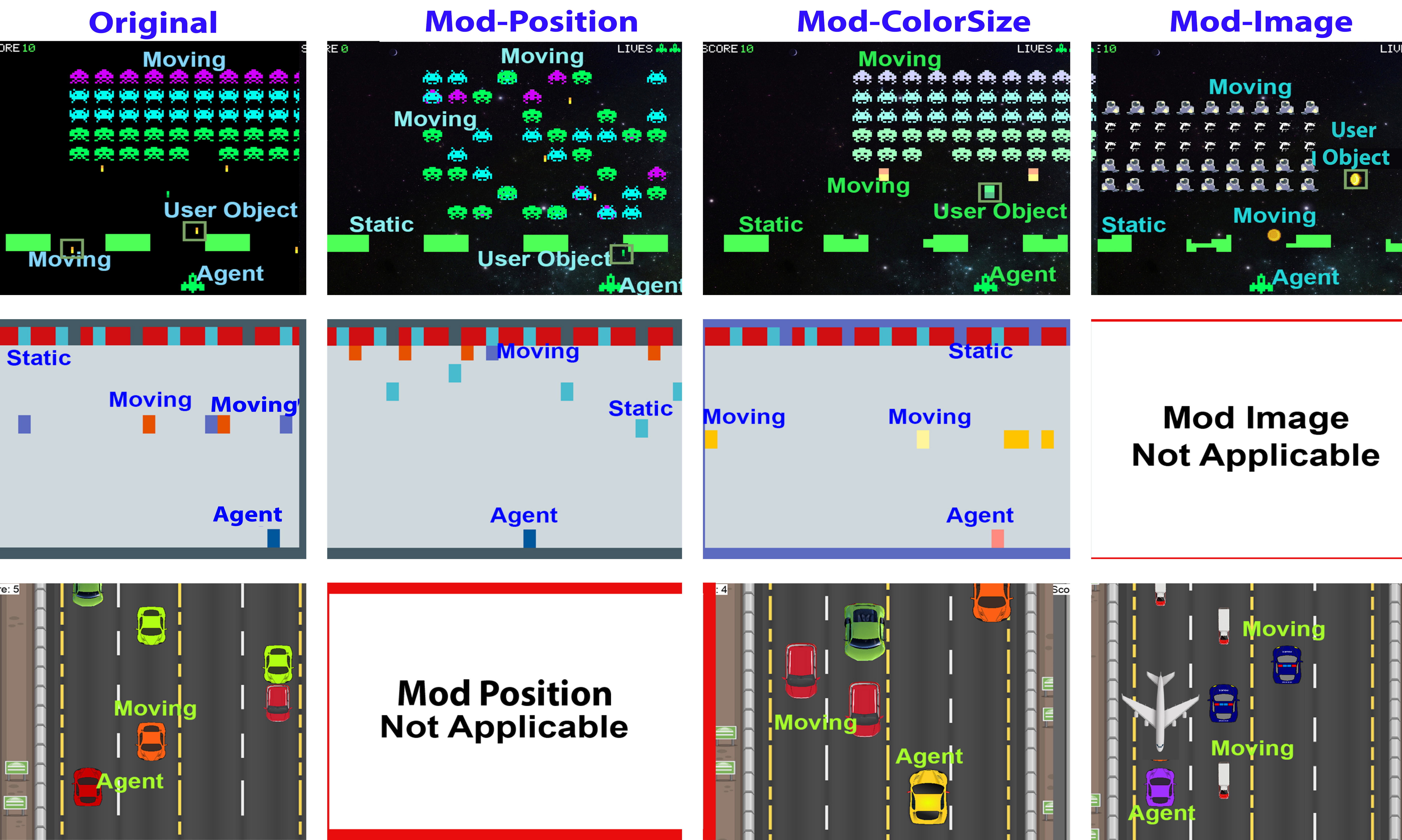}
    \vspace{-2.5mm}

    \caption{From Left to Right, each column shows the original games, Position Modification, Color and Size Modification, and Image Modification. For GVGAI games, all objects are rectangles with fixed shapes and sizes, and Image modification is not applicable. For Roadrash, enemy cars are randomly spread over the road, and thus Position Modification is not applicable.  }
\label{fig:gamesMod}
\end{figure*}

\subsection{Categories}
Theory-based RL methods, even if showing human-like learning traits, use an object-interaction definition known a priori and focus on exploration and planning with a very strong world model~\cite{tsividis2021human}. We take a different route here. Rather than working with the objects directly, we focus on affordance-based object categories inspired from intuitive physics. %Such affordances are learned either from the past or the current game experience. 
These categories are motivated by perceptual signals such as identifying objects as static or moving and then the agent learns their attributes from experience. Humans also learn such object categories having similar affordances over isolated entities and tend to generalize strategies from previously learned knowledge to unseen situations~\cite{perfors2009learning,medin1987family}.

% \cite{medin1987family} humans learn categories over isolated entities. they tend to generalize strategies  that is they are able to map previously learned strategies to unseen situations.\cite{tomov2021multi}}

For all our games, we utilize only these five simple categories:% motivated by the affordances they could provide:
\begin{itemize}
    \item \textbf{Agent} - Agent detection relies on a minimal set of inductive biases, which varies depending on the complexity of the environment.

    \item \textbf{Static objects} - The positions of these objects remain unchanged in successive frames. In simple games, they could be harmless, offering secondary benefits like protection from bullets. In a more complicated setting, they could be part of a winning precondition. 

    \item \textbf{Moving-Good objects} - These objects change their previously occupied positions. They are the interesting and primary interacting entities apart from the agent and may give a positive reward for touching. 
   
    \item \textbf{Moving-Bad objects} - This other moving category represents the prime obstacles in the game. They give negative rewards or kill the agent on touching. As we perceive a threat and move away, the primary affordance associated with this class is to avoid them.
    \item \textbf{Agent objects} - Primarily bullets spawned by the agent.
   
After a player learns these categories, downstream classification becomes instantaneous. Similar to humans, we store their characteristic properties, such as color and classify them instantaneously as they become visible on the screen. %nce the colors are identified, new objects can beassigned to their respective categories. 
%    Once these representative properties like color and shapes are learned, we can automatically categorize existing objects without further ado.
    
\end{itemize}

\section{Identifying the Agent}
As discussed in the previous section, of all the object categories, the agent is of primal importance and requires special attention. Based on previous studies on infants~\cite{spelke1990principles,spelke2007core}, we utilize a set of inductive biases to mimic how a new player would detect the agent in the game.

\subsection{Inductive biases}
Even though identifying objects and the associated properties occurs concurrently and continuously, we try to solve the agent identification problem by utilizing as little information as possible. Thus, we use a sequence of inductive biases for agent identification, stopping at whichever one yields a unique agent representation. In order of priority, these biases are:

\textbf{Inductive Bias 1 - Uniqueness.
}
%Image example
This property suggests that the agent is expected to have a unique form. On the game screen, if two objects appear visually similar, they are less likely to be the agent. % as shown in Fig ref.

\textbf{Inductive Bias 2 - Permanence.}
From a gaming perspective, "permanence" refers to the sustained existence of an entity on the game screen. As the game world is centered around the agent, other objects enter and exit, but the agent is expected to persist at all times.% unless killed.% by an undesirable interaction.

\textbf{Inductive Bias 3 - Action-Object Motion binding.}
The agent is meant for action. As a final conclusive test, we assess all the objects for their mobility with different key presses, the intent being that the agent, as an active principle in the game, would be dynamic rather than passive unless killed by an undesirable interaction. Moreover, as a specific key is pressed repeatedly, only the agent is expected to consistently manifest a repeated action, as outlined by \citeA{de2023self}.

\subsection{Agent Action Key Bindings}
This involves learning the activities an agent does in response to different key presses. It is a form of reinforcement where the player presses keys to observe the agent's behavior. Through repeated iterations of this exercise, the player gradually discerns the mapping of each action to a specific outcome on the screen. We apply a similar principle in our games by taking random actions and observing the changes in the agent's position to map the action-key bindings.
Thus, evaluating movement action key bindings is straightforward. For bullet firing, we check for the generation of a new object near the agent immediately after a keypress. If this occurrence repeats until a specified threshold, we assign the key's action affordance as "Fire."

\section{Implementation}
% There is plenty of literature on object detection and tracking. 
As object detection is a well-researched field, for our tests, we commence with a preexisting list of objects. Subsequently, we categorize these objects into the aforementioned groups solely based on their bounding box and color.
% \textbf{Object representation} We take them as given, learning object representation is already an active area of research (\cite{janner2018reasoning},???) . For example 
%  this paper (https://dspace.mit.edu/bitstream/handle/1721.1/144497/Kapur-shreyask-meng-eecs-2022-thesis.pdf?sequence=1&isAllowed=y)train a neural-object detector and use GNN for tracking. There are various other approaches also like which try to tackle this problem using .....
% but we take them as given to focus on testing the agent ability to learn game playing with learnable categories.

\begin{figure}[tbhp]
    \centering
    \includegraphics[width=\linewidth]{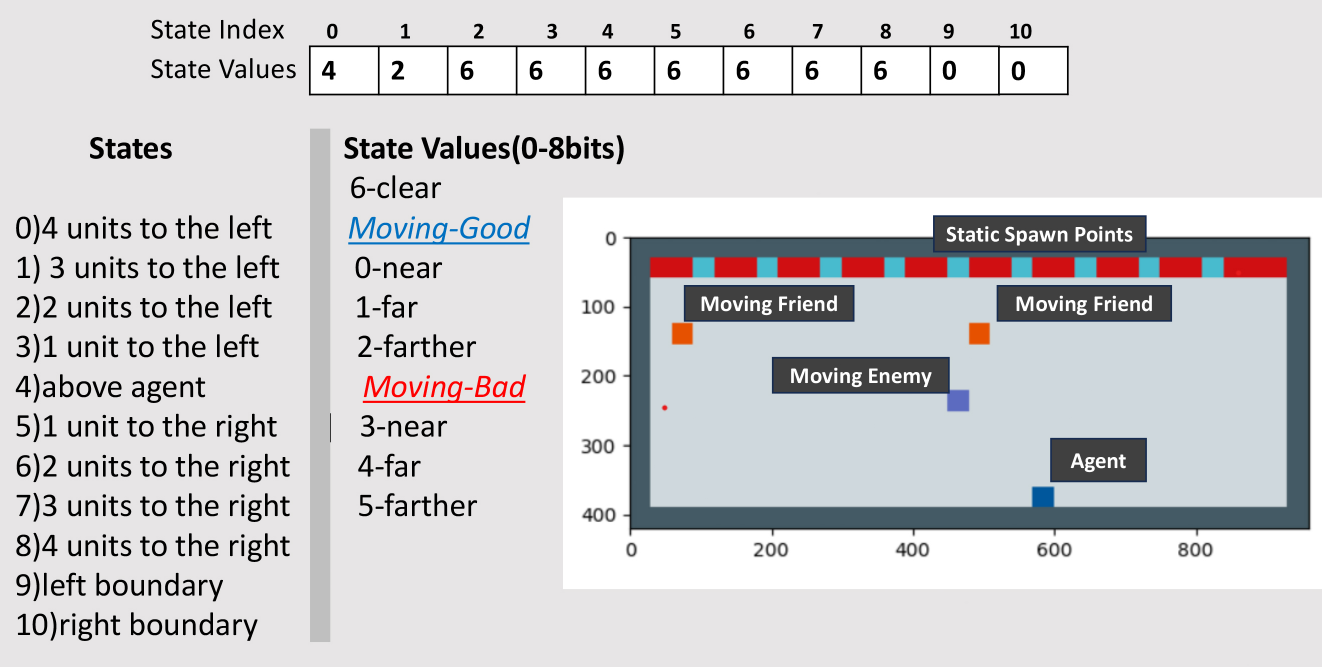}
        \vspace{-2.5mm}

    \caption{MyAliens State Repsentation.}

    \label{fig:MyAliensV1ExampleState}
             \vspace{-2.5mm}

\end{figure}

Incorporating the above object definitions, we try to learn game playing using the Q-learning algorithm \cite{watkins1992q}. For Q-learning to work, we need a state representation that is concise and, at the same time, rich enough for the agent to have sufficient winning information. Consequently, we parse all object category details into state representations tailored to each game setting.
Specifically, we take 2k+1 relative orientation bits, two bits to denote the left and right boundary, and 1 bit to mark the presence of agent bullet if applicable to the game. The 2k+1 orientation bits represent the time it takes for the agent to reach each bit while stationed at the $k^{th}$ bit and store the time it would take for a moving object to cover the horizontal distance from the agent. Here, k varies depending on the requirements of each game (refer Figure \ref{fig:MyAliensV1ExampleState} for an example with k = 4).

\subsection{GVGAI Games}
We modify the MyAliens game from GVGAI framework~\cite{gvgaibook2019} into two variants to test our hypotheses on human-like learning.

\begin{figure*}[htbp]
    \centering
    \includegraphics[width=0.95\linewidth]{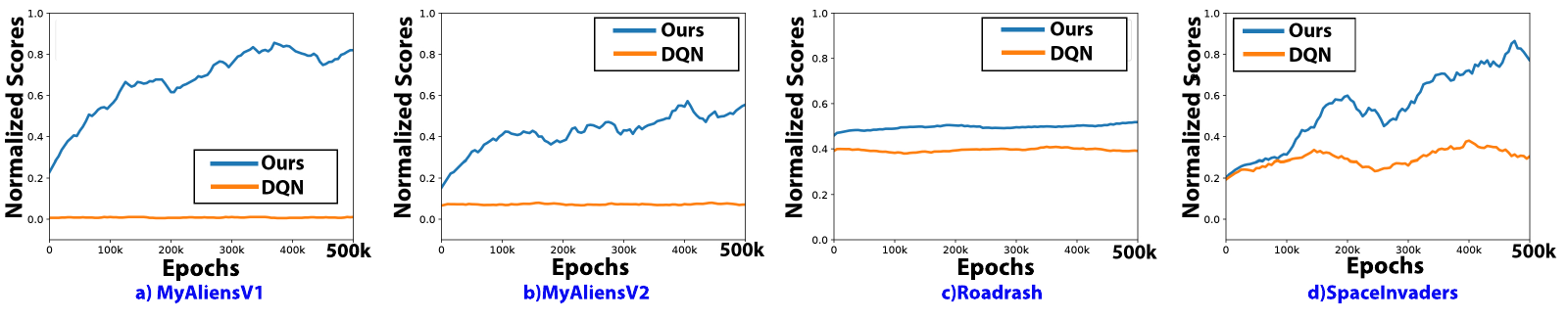}
    \vspace{-2.5mm}
    \caption{Affordance-based Q-learning (Ours) vs. Image-based DQN Normalized Score per epoch plots. a) MyAliensV1: DQN is probably still exploring as it could not learn any meaningful action. b) MyAliensV2 - Both algorithms found difficulty; Q-learning still fairs better, but DQN could not clear even the first level for both variants of MyAliens. c) Roadrash - Very stochastic game with many occasions where avoiding collision is impossible. Q-learning still does better than DQN. d) SpaceInvaders - our algorithm easily learns gameplay using its object-based representation.}
    \label{fig:4Games_epoch}
\end{figure*}

 % DESCRIPTION BRIEF
\subsubsection{MyAliens - variant 1 (MyAliensV1).}
In this game, the objective is to avoid getting hit by any moving object falling from the top till timeout, as they all kill the agent on touching. Game Categories - Agent, Static: Enemy Spawn Points, Moving-Bad: Enemies.

\subsubsection{MyAliens - variant 2 (MyAliensV2).}
This game has two types of moving objects - one food item giving a positive reward and another enemy killing the agent. The agent has to learn to collect ten positively rewarding objects before timeout to win the game. Game Categories - Agent, Static: Enemy Spawn Points, Moving-Bad: Enemies, Moving-Good: Food items (Figure \ref{fig:MyAliensV1ExampleState}).

\subsection{Custom Games}
Additionally, we also test two custom games to check our hypothesis on more visually exciting games.
\subsubsection{Roadrash - Car Driving.}
In this game, the player car has to avoid crashing into the incoming traffic cars. There are only two categories present - agent and moving bad objects. The vehicles can drive only in 4 lanes, making the game very challenging under heavy traffic (Figure \ref{fig:gamesMod}). Game Categories - Agent, Moving-Bad: Enemies.

\subsubsection{SpaceInvaders.}
This game is based on the classic Atari Space Invaders and has the same features with better visuals. The enemies travel horizontally and then move a row down while shooting bullets at the agent spaceship. The agent can shoot only one bullet at a time. Game Categories - Agent, Static: Shields, Moving-Bad: Enemy Spaceships and Bullets, Agent-object: Agent Bullets.

\begin{table*}[tb]
\centering
    \caption{Normalized score for DQN vs. our method averaged over 20 runs of the games. All the models are trained for 1 Million epochs. SpaceInvaders has two levels and a maximum achievable score of 1000. MyAliensV1 has five levels with a maximum score of 50 and MyAliensV2 has three levels with a maximum score of 30. Roadrash ends if the agent can avoid collision for 300 steps, and the score is measured in the number of steps survived. }
        \label{tab:AllResults}
\vskip 0.12in
\begin{small}
  \renewcommand{\arraystretch}{1.1}
  \begin{tabular*}{\linewidth}{@{\extracolsep{\fill}} lcccccccc }

    \toprule
    \multirow{2}{*}{Modifications} &
      % \multicolumn{2}{c}{\textbf{Joint}} &

      % \multicolumn{2}{c}{\textbf{Joint+Vel+BoneVec+BoneAng}} &
      \multicolumn{2}{c}{\textbf{MyAliensV1}} &

    % \multicolumn{2}{c}{\textbf{Joint+BoneVec}} &
    \multicolumn{2}{c}{\textbf{MyAliensV2}} &

      % \multicolumn{2}{c}{\textbf{Joint+Vel+Acc}} \\
      \multicolumn{2}{c}{\textbf{Roadrash}} &

          \multicolumn{2}{c}{\textbf{Space Invaders}} \\

      & {DQN} & {Ours} & {DQN} & {Ours}  & {DQN} & {Ours} & {DQN} & {Ours}  \\
      \midrule
    Random Action     &  \multicolumn{2}{c}{- 0.20}  & \multicolumn{2}{c}{- 0.33}    &  \multicolumn{2}{c}{0.27}    &    \multicolumn{2}{c}{0.27}     \\
    Base-Variant        & - 0.08  & 0.80  &  - 0.23 & 0.57 &  0.45 & 0.50 & 0.51   & 1.0  \\

    Mod-Position    &    - 0.20  & 0.74 &  - 0.27 & 0.52 & NA & NA & 0.28  &  0.42   \\
    Mod-ColorSize         & - 0.20  & 0.80 &  - 0.27 &  0.57 & 0.40 & 0.48 &     0.31  & 1.0  \\
    Mod-Image            &  NA & NA  & NA  &  NA & 0.38 & 0.47 & 0.30  & 1.0   \\
    % Mod-Bullets     &  &   &    &   &   &   &  &  \\
    % Mod-NoBlocker   &  &   &  NA  &  NA & NA  &  NA & NA & NA \\

    \bottomrule
  \end{tabular*}

      \end{small}
\vskip -0.1in
\end{table*}

\section{Generalization Experiments}
Our goal here is not to defeat a Deep Reinforcement learning algorithm but to show that using a methodology like ours has certain benefits that opens up new avenues for mimicking human learning characteristics. For all the tests, we train DQN for 10e6 with linear learning rate decay from 1 to 0.01.

% Test 1
First, we test our games to see their learning capacity compared against a DQN agent for all four games. For this, we plot the normalized average scores over 20 runs for model vs. epoch, where epoch is defined as one game run loop. We normalize the scores as follows:
\[ Normalized\;score =\frac{ actual\;score}{maximum\;achievable\;score}\]
MyAliensV1 has five levels with different placement of spawn points for moving enemies with a maximum score of 50, +10 for winning each level, and -10 for losing. We test MyAliensV2 for three levels, with a maximum achievable score of 30. The agent receives a -10 reward for losing, and a reward of +1 for collecting a food item. The level is won when ten such items are collected.

In Roadrash, the agent needs to survive the traffic onslaught for 300 steps. SpaceInvaders has two levels with a maximum achievable score of 1000, from +10 received for killing each of the 50 enemy spaceships over the two levels.

%Test2
Humans, after learning how to play a game, would easily adapt to slight variants of the game.
To test the generalized ability of our agent to mimic human-like learning, we run the agent in different variations of the games. For all such cases, we train the game only on the base variant of the game.

We explore three types of game variations (Figure \ref{fig:gamesMod}):
\begin{itemize}
    \item \textbf{Mod-Position}: Partially random placement of moving enemies (SpaceInvaders) or static enemy spawn points (MyAliens).
    \item \textbf{Mod-ColorSize}: Alters size and color of game objects.
    \item \textbf{Mod-Image}: Substitutes default game images.
% CAN ADD THESE TWO MODIFICATION BUT ONLY APPLICABLE FOR SPACE INVADERS 
   % \item Mod-Bullets: Increases the number of spawned enemy bullets (Space Invaders only).  
  %  \item Mod-NoBlocker: Removes static bullet blockers (for Space Invaders only)
\end{itemize}

GVGAI games do not permit the modification of object sizes, and all in-game objects are constructed using unit-sized colored rectangles. Thus, the Mod-Image variant is not applicable, as it involves using external images, which the GVGAI framework does not support. In Roadrash, where enemy cars are spawned randomly, the Mod-Position variant will not yield any new variation and is not used.

We train the DQN algorithm using a batch size 32 on each game run loop with experience replay. The Q-learning agent is trained only once on each run on the latest experience. Thus, even on the same level of epochs, DQN weights are updated 32 times more than Q-learning. 

\section{Observations}

To test the efficacy of our category-level representations, we run two kinds of tests. First, we compare DQN and our method under varying training durations. Scores from the trained models from different training epochs are plotted, and we analyze the agent's normalized score (Figure~\ref{fig:4Games_epoch}).
 One Epoch is defined as one run of the game loop. Even though Q-learning updates 32 times less than DQN, it is able to learn correct decisions quickly. We plot the results up to 0.5 million epochs, equivalent to approximately 2 hours of gameplay at 60 frames per second, and in most cases, DQN did not show any improvement owing to its sample inefficiency. 

Our algorithm demonstrates strong performance in MyAliensV1, successfully winning all five levels. In SpaceInvaders also, it wins both the levels. However, MyAliensV2 presents a more challenging scenario, requiring the agent to distinguish between moving-good and moving-bad categories and collect ten of the good ones before a time-out to win the level.
Here also, our method does well, but the performance degrades as compared to MyAliensV1, primarily because our agent has only a 9-bit state space representation, i.e., it can see nearest objects only within a range of four units on both the left and right sides. 
Given that the moving-good category is dispersed over a broader x-range of thirty bits, the agent often struggles to locate the moving-good category within its narrow field of vision. Consequently, the timer runs out before the agent can collect the required ten items, impacting the overall score.

 The escalating difficulty in subsequent levels, coupled with a reduction in the number of Moving-Good spawn points, adds to the complexity of the task.
We also tested a broader state representation of 25 bits, but the learning became computationally intractable, and the agent struggled to learn meaningful affordances. Nevertheless, even with a limited view, our agent clears the first two levels and fails only on the final level (Table \ref{tab:AllResults}).

For Roadrash, even though our method does better, we do not see substantial performance gain with training. The game has four lanes, with enemies spawning stochastically in any of them.
Thus, in many cases, all four lanes get blocked, and a crash becomes unavoidable. In other situations, avoiding accidents requires precise control because of the crowded structure of game objects. So, even a reasonably learned agent could not perform well in this game, and the performance was more-or-less stagnant. Nonetheless, our algorithm still fairs better against the DQN agent.

Our second set of comparisons focuses on the transferability of the acquired knowledge.
For deep learning algorithms, object level alterations, such as changing object colors, can have devastating consequences \cite{lake2017building}. On the other hand, humans can easily manage such variations.
Our results indicate that, unlike a DQN agent, our category-based method exhibits similar performance scores, aligning more with human-level gameplay (Table \ref{tab:AllResults}).

This is primarily because, at the category level, the state representations remain relatively stable despite the aforementioned generalization modifications. Consequently, our algorithm's performance does not degrade with these variations.
It's noteworthy that both models in these comparisons are trained for one million epochs. However, for MyAliensV1 and MyAliensV2, DQN is still in its exploration phase, exhibiting minimal performance improvement, and the introduced variations further degrade its performance. 
This is particularly evident in SpaceInvaders, where the DQN agent while displaying some learning traces in the base variant, regresses to the level of a random agent when faced with varying input pixel combinations. 
As the Roadrash game is challenging from the start, there is little difference after making a difficult game more difficult.

Among all the alterations, only SpaceInvaders Mod-Position resulted in a substantial decline in Q-learning performance. This is primarily due to position modifications creating new, and previously unseen, state representations.
In this setting, as the enemies get randomly arranged, some enemies get placed too close to the agent. As such states are previously unseen, a table-based Q-learning agent struggles to navigate this variation (Table~\ref{tab:AllResults}). 
Such instances could potentially be avoided by using techniques to extrapolate for unseen states based on prior experience. Apart from this, other game modifications consistently exhibit performance similar to the unmodified original versions of the games, as is also expected from a human player.

%Variations where we can test sort of genralization where model free algorithms failed. To the best of our knowledge we don't find such generalization testing with environment perturbations in proper works which advocates the use of category level object representation which is the major contribution of this work.

Thus, our comparisons show that an object-based representation, even if applied within a model-free framework, offers much better sample efficiency (Figure \ref{fig:4Games_epoch}).
This improvement is evident in results with environmental perturbations, such as varying enemy positions and differently shaped enemies, among other variations. 
The primary factor contributing to this enhanced performance is the category-based representation, in which minor perturbations do not alter the game representation significantly, while causing problems in approaches where objects are treated as separate entities, and also for pixel-based model-free methods like DQN.

\section{Conclusion}
Making machines learn and act like humans is an important goal in Artificial General Intelligence(AGI)~\cite{lake2017building,tsividis2017human,pouncy2022theory}. In this paper, we look at video game playing from the eyes of a novice player discovering gameplay dynamics. Drawing inspiration from Spelke's conception of core physics knowledge~\cite{spelke1990principles,spelke2007core}, we developed object category representations that transfer well across simple games. Building upon this state representation, we show that machines can exhibit certain similarities to human-like learning in game playing. In contrast with existing approaches that learn to play games using self-learned pixel-based representations~\cite{mnih2015human,hessel2018rainbow}, our approach focuses on using composable affordance-based object representations and shows faster and more robust learning in such games than is seen in foundationally pixel-based approaches. 

It is natural, of course, to question the degree to which intuitive physics priors handcrafted to align with game-world task requirements generalize to more natural settings. While answering such questions is a longer-term project, we point out that the semiotics and affordances of video games, particularly classic games of the sort we test, have been optimized for easy comprehension across language and age barriers~\cite{blomberg2018semiotics}. Thus, the success of classic intuitive physics heuristics~\cite{spelke1990principles} in producing affordance-based representations that generalize across games may well be because these heuristics apply well in the real world also. 

% \textbf{
% \cite{perfors2009learning} - Learning to categorize objects in the world is more than just
% learning the specific facts that characterize individual categories. We can also learn more abstract knowledge about how
% categories in a domain tend to be organized – extending even
% to categories that we’ve never seen examples of. These abstractions allow us to learn and generalize examples of new
% categories much more quickly than if we had to start from
% scratch with each category encountered.}

Due to its model-free nature, our agent is still not as sample-efficient as a human, but it does well on the generalizability task. Theory-based RL approaches take the model as given and explore planning within such a framework~\cite{tsividis2021human,pouncy2022theory}. Future work may extend the use of affordance-based object representations to learn a model of the environment based jointly on core knowledge and experience.

\bibliographystyle{apacite}

\setlength{\bibleftmargin}{.125in}
\setlength{\bibindent}{-\bibleftmargin}

\bibliography{CogSci_Template}

\end{document}